\documentclass[letterpaper]{article} 
\usepackage{EquiBoost}  
\usepackage{times}  
\usepackage{helvet}  
\usepackage{courier}  
\usepackage[hyphens]{url}  
\usepackage{graphicx} 
\usepackage[backend=bibtex,
style=authoryear]{biblatex} 
\usepackage{caption} 
\usepackage{algorithm}
\usepackage{algpseudocode}

\usepackage{booktabs}
\usepackage{multirow}
\usepackage{amsmath}
\usepackage{newfloat}
\usepackage{listings}
\DeclareCaptionStyle{ruled}{labelfont=normalfont,labelsep=colon,strut=off} 
\floatstyle{ruled}
\newfloat{listing}{tb}{lst}{}
\floatname{listing}{Listing}

\title{EquiBoost: An Equivariant Boosting Approach to Molecular Conformation Generation}
\author {
    Yixuan Yang\textsuperscript{\rm 1},
    Xingyu Fang\textsuperscript{\rm 1},
    Zhaowen Cheng\textsuperscript{\rm 1} \\
    Pengju Yan\textsuperscript{\rm 1}\thanks{Corresponding Authors},
    Xiaolin Li\textsuperscript{\rm 1}\footnotemark[1]
}
\affiliations {
    \textsuperscript{\rm 1}HIM, Chinese Academy of Sciences\\
    \{yangyixuan, fangxingyu, chengzhaowen\}@him.cas.cn \\
    yanpengju@gmail.com, xiaolinli@ieee.org
}

\usepackage{bibentry}

\usepackage{amssymb}

\begin{document}

\maketitle

\begin{abstract}
Molecular conformation generation plays key roles in computational drug design. Recently developed deep learning methods, particularly diffusion models have reached competitive performance over traditional cheminformatical approaches. However, these methods are often time-consuming or require extra support from traditional methods. We propose EquiBoost, a boosting model that stacks several equivariant graph transformers as weak learners, to iteratively refine 3D conformations of molecules. Without relying on diffusion techniques, EquiBoost balances accuracy and efficiency more effectively than diffusion-based methods. Notably, compared to the previous state-of-the-art diffusion method, EquiBoost improves generation quality and preserves diversity, achieving considerably better precision of Average Minimum RMSD (AMR) on the GEOM datasets. This work rejuvenates boosting and sheds light on its potential to be a robust alternative to diffusion models in certain scenarios.
\end{abstract}

\begin{links}
    \link{Code}{https://github.com/foralan/EquiBoost}
\end{links}

\section{Introduction}

The three-dimensional structure of a molecule, known as its conformation, significantly influences molecular properties, such as free energy, chemical reactivity, and biological activity \parencite{wu2018moleculenet,li2015importance}. Consequently, molecular conformation generation is a fundamental task in computational drug design \parencite{hawkins2017conformation}, impacting a wide range of applications from virtual screening \parencite{kuan2023keeping} to molecular docking \parencite{jiang2024pocketflow} and molecule binding affinity predictions \parencite{wang2024structure}. To generate molecular conformations, both cheminformatics and deep learning methods have been employed.

Cheminformatics methods often involve template searching and energy minimization. These template-based methods face significant challenges due to the combinatorial explosion of conformations as the number of rotatable bonds increases, coupled with incomplete sampling of the conformational space \parencite{zhang2022optimization}. Additionally, the low accuracy of force fields can introduce biases, particularly when dealing with unreasonable initial input conformations \parencite{zhou2023deep,hawkins2017conformation}. 

Deep learning methods, on the other hand, enable atom-level research, eliminating the dependence on fixed templates. With the advancement and success of diffusion models in image generation \parencite{rombach2022high, ramesh2022hierarchical}, these approaches have also been adopted in chemistry \parencite{hoogeboom2022equivariant, abramson2024accurate, song2024equivariant}, including molecular conformation generation. While diffusion models have shown promising performance, they are not without limitations. Due to their iterative nature, requiring thousands of steps to generate outputs, diffusion models often suffer from inefficiencies in sampling. In the context of molecular conformation generation, some diffusion model variants have reduced the number of sampling steps from thousands to tens or even a single step, but this often comes at the expense of accuracy, leading to suboptimal results \parencite{jing2022torsional, fan2023ec}.

In this work, our key contributions are as follows:
\begin{itemize}
\item \textbf{A novel equivariant boosting method}: We introduce EquiBoost, a boosting model that sequentially integrates multiple equivariant graph transformers as weak learners to iteratively refine 3D molecular conformations. During training, EquiBoost rapidly converges to high accuracy. In addition, EquiBoost also improves efficiency by reducing the number of inference steps from thousands in diffusion models to just five.
\item \textbf{Significant performance improvement on the GEOM dataset}: EquiBoost surpasses previous methods by enhancing generation quality while maintaining diversity on the GEOM datasets. These results indicate the potential of EquiBoost as a robust alternative to diffusion models for molecular conformation generation.
\end{itemize}

\section{Background}
\subsection{SE(3) equivariance}

Equivariance describes a property of a function where a transformation applied to the input results in a corresponding transformation of the output \parencite{scott2012group}. Formally, for a given transformation $g$ and function $f$, the function is equivariant if $f(g{\cdot}x)=g{\cdot}f(x)$. 

Molecular structures in 3D space exhibit rotational and translational invariance in their physical and chemical properties. Moreover, the coordinates of atoms change with rotation and translation transformations. Therefore, in the task of conformation generation, our focus lies on the Special Euclidean group in three dimensions, SE(3), with the goal of achieving SE(3) equivariance. 

\subsection{Equivariant graph neural networks}
Graph neural networks (GNNs) belong to a class of artificial neural networks operated on graph-structured data \parencite{wu2022graph}. They have been widely applied in various fields, including chemistry, social networks and recommendation systems. In the realm of chemistry, molecules can be naturally represented as graph structures, where atoms correspond to nodes and chemical bonds to edges \parencite{sanchez2021gentle}. By propagating information through nodes and edges, GNNs effectively learn representations from molecules. However, traditional GNNs are not capable of handling common transformations such as rotation and translation. Adding equivariance into GNNs addresses this limitation.

The introduction of Equivariant Graph Neural Networks (EGNNs) \parencite{satorras2021n} marks a significant advancement in this area. A major distinction is the update rule for node positions, where positions are adjusted using a weighted sum of relative distances. This design ensure the model's equivariance to the Euclidean group $E(3)$. Some other works address the equivariance requirements using principles from group theory \parencite{geiger2022e3nn}. These approaches often utilize spherical harmonics to convert common features to equivariant ones and employ tensor product to ensure equivariance within the neural network \parencite{thomas2018tensor,fuchs2020se,liao2023equiformer}. 

\subsection{Diffusion models}
Diffusion models are a type of generative models containing a forward process and a reverse process. The forward process involves gradually adding Gaussian noise to a initial datapoint $x_0$ until it becomes a purely white noise datapoint $x_T$ in standard Gaussian distributions. Given $x_0\sim p(x_0)$ and $x_T\sim p(x_T)$, the forward process can be described by the following Markov chain:
\begin{equation}
q(x_{1:T} | x_0) = \prod_{t=1}^T q(x_t | x_{t-1})
\end{equation}
\begin{equation}
q(x_t | x_{t-1}) = \mathcal{N}(x_t; \sqrt{1 - \beta_t} x_{t-1}, \beta_t \mathbf{I})
\end{equation}
where $\beta_t$ is the fixed variance schedule.
The reverse process aims to recover the original data from the prior distribution:
\begin{equation}
p_\theta(x_{0:T}| x_t) = p(x_T )\prod_{t=1}^T p_\theta(x_{t-1} | x_t)
\end{equation}
\begin{equation}
p_\theta(x_{t-1} | x_t) = \mathcal{N}(x_{t-1}; \mu_\theta(x_t, t), \sigma^2 \mathbf{I} )
\end{equation}
where $p_\theta(\mathbf{x}_{t-1} |\mathbf{x}_t)$ is the learnable Markov transition kernel, a.k.a, the denoising model. $\mu_\theta$ are parameterized neural networks to estimate the means, $\sigma$ is the custom variance.

\begin{figure*}[t]
    \centering
    \includegraphics[width=0.9\textwidth]{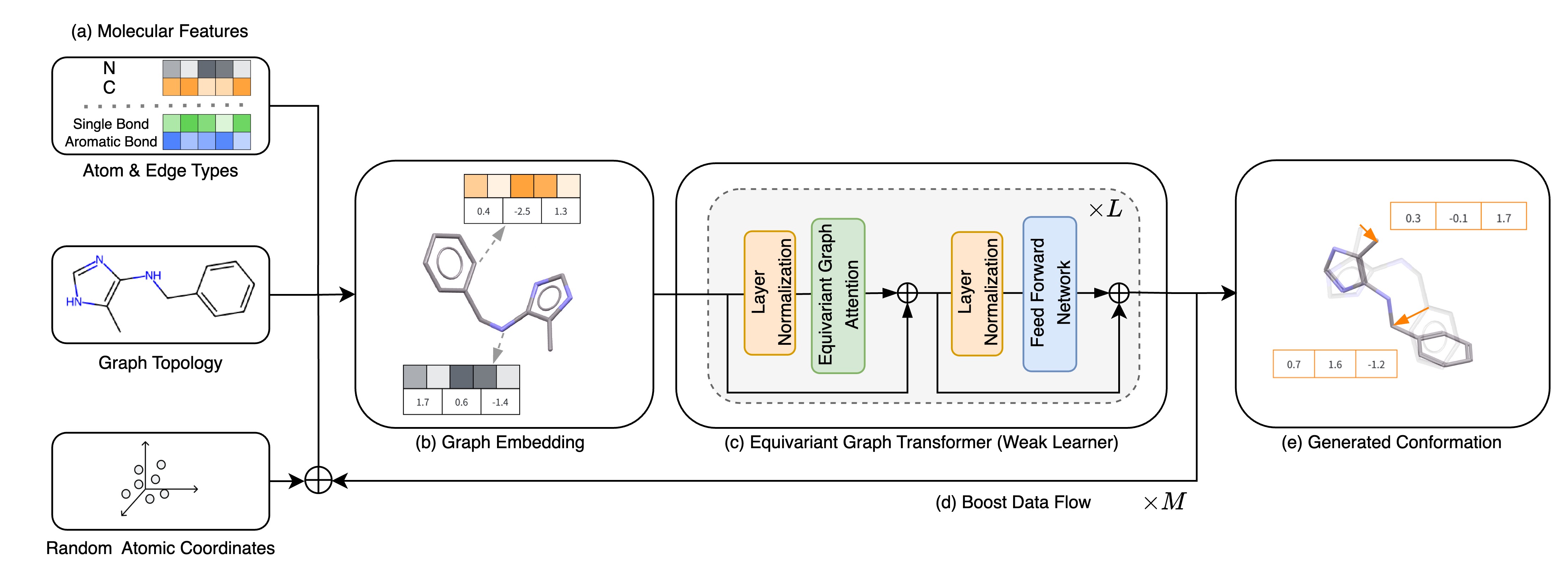}
    \caption{\textbf{EquiBoost framework} (a) The molecular features include three components: the atom and bond types, graph topology, and random atomic coordinates. (b) The graph embedding is constructed from these three molecular features. (c) The equivariant graph transformer contains $L$ blocks. Each block contains equivariant graph attention, layer normalization, and feed-forward network. This transformer model serves as the weak learner. (d) In the boosting paradigm, the generated conformation from the current weak learner is passed to the next, and this process is repeated for $M$ iterations. (e) The final conformation is generated after $M$ boosting iterations. The orange arrow indicates the output of the weak learner, which represents the displacement of atomic coordinates.}
    \label{fig1}
\end{figure*}

\subsection{Gradient boosting}
Boosting refers to combining several base (weak) learners into a strong model to generate more accurate results. Each weak learner is trained sequentially, correcting the output from the previous model \parencite{freund1997decision,chen2016xgboost,ke2017lightgbm}. Gradient boosting is a special form of boosting that trains the model by gradually minimizing the loss function using gradient descent. It is considered more robust and extends the application field to multi-class classification and regression \parencite{natekin2013gradient}.

The gradient boosting process can be summarized mathematically as follows:

Given the training set $\{(x_{i},y_{i})\}_{i=1}^{n}$, where $x$ is the input variable and $y$ is the output variable. The process starts with an initial model $F_0(x)$, typically a constant value that minimizes a differentiable loss function $L$.

Next, for iteration $m$ from $1$ to $M$, the negative gradient for the current model is computed:
\begin{equation}
    r_{m} = -\left[ \frac{\partial L(y, F(x))}{\partial F(x)} \right]_{F(x) = F_{m-1}(x)}
\label{eq:residue_define}
\end{equation}
followed by fitting a new weak learner $h_m(x)$ to these negative gradients $r_m$. The model is subsequently updated by incorporating the new learner:
\begin{equation}
    F_m(x)=F_{m-1}(x)+h_m(x)
\end{equation}
After $M$ iterations, the final model is:

\begin{equation}
F_M(x) = F_0(x) + \sum_{m=1}^{M} h_m(x)
\label{eq:gb_define}
\end{equation}

In the gradient boosting process, each weak learner $h_m(x)$ is trained to correct the errors of the preceding model $F_{m-1}(x)$ by focusing on the gradient of the loss function. The final prediction is obtained by the predictions from all weak learners.

\section{Related work}
Before the rapid advancement of deep learning, two primary strategies are commonly used for generating molecular conformations: the stochastic approach and the systematic approach \parencite{riniker2015better}. Stochastic approaches, primarily based on Molecular Dynamics or Markov   chain Monte Carlo (MCMC) techniques, are known for being computationally slow \parencite{de2016role,gilabert2018monte,bai2022application}. The systematic approach, employed by software like OMEGA \parencite{hawkins2010conformer,hawkins2012conformer}, relies on predefined libraries of common torsion angles which limits its applicability to complex molecules. These two strategies are combined in the ETKDG method, which typically begins by initializing conformations using distance geometry and then fine-tuning them with a force field \parencite{wang2020improving}. However, this two-stage process can lead to error accumulation and remains time-consuming. Since the release of RDKit 2018.09, ETKDG has become the default method for conformation generation \parencite{landrum2013rdkit}.

In recent years, several deep learning models have been proposed for molecular conformation generation \parencite{guan2021energy,alakhdar2024diffusion,xu2024geometric,peng2023moldiff}. Among them, GeoMol \parencite{ganea2021geomol} is an end-to-end SE(3)-invariant model that jointly predicts torsion angles and local structures using message passing neural networks (MPNNs) and self-attention networks. GeoMol outperforms existing machine learning methods like GraphDG \parencite{simm2020generative} and traditional methods such as RDKit. Another notable model is GeoDiff \parencite{xu2022geodiff}, a generative model recognized for its strong performance in Euclidean diffusion approaches for conformation generation, demonstrating superior accuracy on various benchmarks. However, its iterative diffusion process is computationally expensive. To address the computational expense of GeoDiff, EC-Conf \parencite{fan2023ec} was proposed, which reduces the denoising step to one, but at the cost of significantly lower accuracy. Another approach, Torsional Diffusion \parencite{jing2022torsional}, models the diffusion process on the hypertorus, where all degrees of freedom except for the torsion angles are fixed. This method utilizes two orders of magnitude fewer denoising steps than GeoDiff. However, model accuracy is still compromised.

\section{Method}

In this section, we elaborate on the framework of EquiBoost. We begin by defining the molecular conformation generation task. Next, we describe the construction of an equivariant model tailored for this specific task and present our theory of equivariant gradient boosting. Finally, we illustrate the objective function by detailing its formulation.

\subsection{Definition of molecular conformation generation}
A molecule is represented as a graph $\mathcal{G}=(\mathcal{V},\mathcal{E})$, where the atoms are nodes $v \in \mathcal{V}$ and the bonds are edges $e \in \mathcal{E}$. The conformation of a molecule is described by the Cartesian coordinates of its atoms, each corresponded to a coordinate vector $c \in {R}^3$ in the 3-dimensional Euclidean space. In this way, the conformation with $N$ atoms can be represented as a matrix $\mathcal{C}=[c_1, c_2, ..., c_N] \in {R}^{N\times3}$. Our goal is to learn a generative model $\gamma_{\theta}(\mathcal{C}|\mathcal{G})$ that samples conformations of a molecule conditioned on its graph $\mathcal{G}$. $\theta$ refers to the model parameters.

\subsection{Equivariant models}

We construct a generative model that directly samples conformations from a random distribution $\mathcal{C}_{noise} \sim \mathcal{N}(0, \mathbf{I}_{N})$, conditioned on its graph representation $\mathcal{G}$, which encodes atom types and bond types. The model is formulated as $\gamma_\theta(\mathcal{G}, \mathcal{C}_{noise})$.

To ensure that the generated conformation remains equivariant to SE(3) transformations applied to the input random coordinates, the model's output should represent the displacement of these coordinates, rendering them SE(3)-equivariant vectors \parencite{watson_novo_2023}. Consequently, the model itself must also possess the SE(3) equivariance property:

\begin{equation}
\forall{g} \in \text{SE(3)},~\gamma_{\theta}(\mathcal{G},g\cdot\mathcal{C}_{noise})=g\cdot\gamma_{\theta}(\mathcal{G},\mathcal{C}_{noise})
\end{equation}

We employ EquiformerV2 \parencite{liao2023equiformerv2}, an equivariant graph attention transformer for 3D atomic graphs, to construct the generative model. The model consists of multiple blocks, with the core components being the equivariant graph attention (EGA) mechanism.

Suppose the equivariant graph transformer has $L$ layers. From $x^L$, the output of the final layer, we extract 3-dimensional vectors $\Delta{\mathcal{C}}=\{\Delta{c_1}, \Delta{c_2}, \cdots, \Delta{c_N}\}$, representing the changes in atom coordinates from the initial noisy conformation $\mathcal{C}_{noise}$. The conformation $\hat{\mathcal{C}}$ is then obtained by adding $\Delta{\mathcal{C}}$ to $\mathcal{C}_{noise}$. This process can be mathematically expressed as follows:

\begin{equation}
    \Delta{\mathcal{C}}=\gamma_\theta(\mathcal{G}, \mathcal{C}_{noise})
\end{equation}

\begin{equation}
    \hat{\mathcal{C}}=\mathcal{C}_{noise}+\Delta\mathcal{C}=\mathcal{C}_{noise}+\gamma_{\theta}(\mathcal{G},\mathcal{C}_{noise})
\end{equation}

For brevity, we omit the details of the equivariance proof and the EGA mechanism. For more information, refer to the Appendix.

\subsection{Equivariant gradient boosting}

\subsubsection{EquiBoost framework} Treating the aforementioned equivariant model as a weak learner, we propose a boosting model that stacks several weak learners, as illustrated in Figure \ref{fig1}. To apply gradient boosting methods while preserving the equivariance property, the conformation generated by one weak learner is fed into the next. The boosting model iteratively refines the conformation through its sequence of weak learners, ultimately producing an accurate conformation. Defining $M$ as the number of weak learners, or steps. One step of EquiBoost is defined as:

\begin{equation}
\mathcal{C}^{m+1}=\mathcal{C}^m+\gamma_{\theta,m}(\mathcal{G}, \mathcal{C}^m)
\end{equation}

To obtain the final generated conformation $\mathcal{C}^M$, we repeat this process as follows:

\begin{align}
\mathcal{C}^M =&\mathcal{C}^{M-1}+\gamma_{\theta,{M-1}}(\mathcal{G},\mathcal{C}^{M-1}) \notag \\
=&\mathcal{C}^{M-2}+\gamma_{\theta,{M-2}}(\mathcal{G},\mathcal{C}^{M-2})~+\notag \\
&\mathcal{C}^{M-1}+\gamma_{\theta,{M-1}}(\mathcal{G},\mathcal{C}^{M-1}) \notag \\
=&\cdots \notag \\
=&\mathcal{C}^{0}+\gamma_{\theta,0}(\mathcal{G},\mathcal{C}^{0})+\cdots+\gamma_{\theta,{M-1}}(\mathcal{G},\mathcal{C}^{M-1}) \notag \\
=&\mathcal{C}^{0}+\sum_{m=0}^{M-1}{\gamma_{\theta,m}(\mathcal{G},\mathcal{C}^m)}
\end{align}
where $\mathcal{C}^0$ is identical to $\mathcal{C}_{noise}$. Given EquiBoost as $\mathcal{D}$, we finally define the model as:

\begin{equation}
    \mathcal{D}(\mathcal{C}|\mathcal{G},\mathcal{C}_{noise})=\mathcal{C}_{noise}+\sum_{m=0}^{M-1}{\gamma_{\theta,m}(\mathcal{G},\mathcal{C}^m)}
\end{equation}

\begin{figure}[!b]
    \centering\includegraphics[width=0.47\textwidth]
    {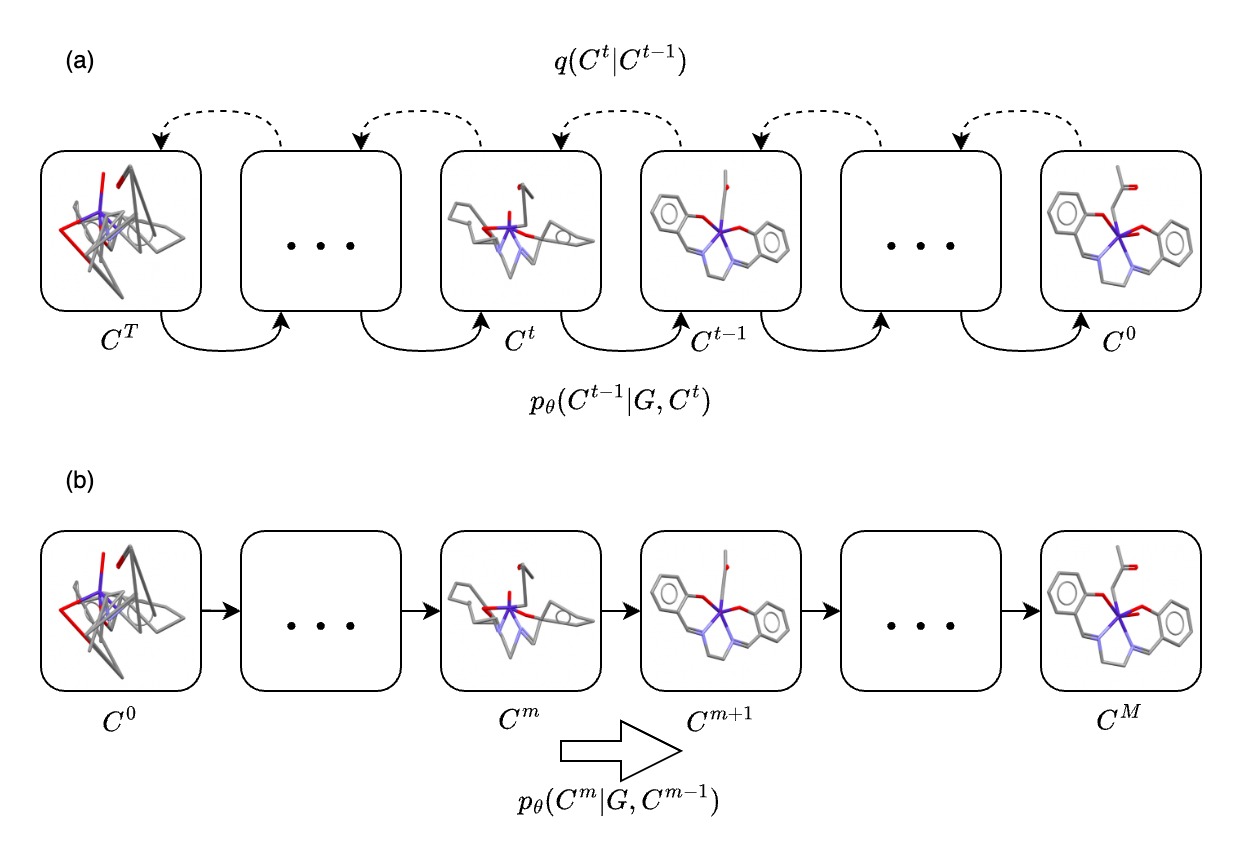}
    \caption{\textbf{Comparison between diffusion models and EquiBoost} (a) A diffusion model contains a forward and reverse process. The forward process, depicted from right to left, involves the gradual addition of noise to the true conformation $\mathcal{C}^0$, resulting in the chaotic conformation $\mathcal{C}^T$. The reverse process, shown from left to right, involves the iterative removal of noise to recover the true conformation $\mathcal{C}^0$. (b) EquiBoost aims to directly generate a precise conformation $\mathcal{C}^M$ from the chaotic conformation $\mathcal{C}^0$.}
    \label{fig2}
\end{figure}

When compared with diffusion models, as illustrated in Figure \ref{fig2}, a commonality lies in the need to sample stable conformations from random noise. The theory behind diffusion models shares a similar design philosophy with gradient boosting: iteratively “rectifying” the prediction. Unlike diffusion models, which involve adding noise to generate a diffused conformation and then learning to recover the denoised state, EquiBoost directly learns to generate precise conformations from noise. The idea behind our method bears resemblance to the recycling mechanism in AlphaFold2 \parencite{jumper2021highly}. The key difference is that while the recycling mechanism refines both the target and the features, EquiBoost refines only the target, keeping the input features unchanged.

\subsubsection{Training strategies} Training EquiBoost directly is time-consuming, as it requires $M$ times more model parameters and training time compared to models without boosting. To accelerate this process, we employ two strategies: weight sharing among weak learners and randomization of the boosting number. First, in the boosting model, all weak learners share a common set of parameters, reducing the total number of parameters from $M$ times that of a single weak learner to that of one weak learner. To distinguish different weak learners, we inject current step ${m}$ as an input feature into the corresponding weak learner. The equation is then transformed to:

\begin{equation}
    \mathcal{D}(\mathcal{C}|\mathcal{G},\mathcal{C}_{noise})=\mathcal{C}_{noise}+\sum_{m=0}^{M-1}{\gamma_{\theta}(\mathcal{G},\mathcal{C}^m, m)}
\end{equation}

This strategy not only cuts down on the model's complexity but also enables EquiBoost to account for the refining steps in addition to the input coordinates. Second, the number of boosting iterations is randomized during training but keeps fixed during testing, which is adopted from \parencite{jumper2021highly}. Specifically, we sample an integer $M_{train}$ from a uniform distribution from 0 to $M-1$, significantly reducing the training time per epoch.

\subsubsection{Initialize conformations with constrained randomization}
We employ a constrained randomization method to initialize noisy molecular conformations, which serves as an alternative to directly sampling all atomic coordinates from random noise. This method was first proposed in Torsional Diffusion \parencite{jing2022torsional}. The underlying theory posits that molecular conformations can be viewed as a combination of intrinsic coordinates and torsional angles. Intrinsic coordinates, including bond lengths, bond angles, and ring conformations, can be easily predicted using classical cheminformatics tools. However, torsional angles, which include the dihedrals of freely rotatable bonds, remain a challenging prediction task.

In Torsional Diffusion, molecular conformations are initialized using RDKit, with random assignment of torsional angles. During optimization, intrinsic coordinates are fixed while torsional angles are iteratively optimized. Compared to directly sampling all atomic coordinates from random noise, this approach leverages prior knowledge embedded in cheminformatics tools. In contrast to Torsional Diffusion, our method directly updates atomic coordinates following constrained randomization. That is, we allow both intrinsic coordinates and torsional angles to change simultaneously during the generation process. Our experimental results confirm the advantages of this method, which we refer to as Constrained Random Sampling (CRS). A detailed comparison between CRS and Random Sampling (RS) can be found in the appendix.

\subsubsection{Optimal conformation mapping} A molecule may possess multiple conformations. When the conformation ensemble is large, predicting an ensemble of conformations of the same size substantially increases the computational expense associated with training.



For diffusion models, aligning predictions with reference conformations is straightforward. The model's input is generated by directly adding noise to a conformation, creating a natural one-to-one mapping between predictions and references. In contrast, the input to EquiBoost during training is random noise, which lacks direct correspondence to reference conformations.

To address this problem, we propose a method called optimal conformation mapping, inspired by the classic optimal transport strategy \parencite{tong2023improving}\parencite{ganea2021geomol}. While optimal transport typically identifies a mapping between two sets of ${K}$ elements that minimizes overall cost, our approach focuses on mapping a single element to a set, minimizing the cost between that element and each member of the set. Specifically, we calculate the root mean square deviation (RMSD) between the initial noisy conformation $\mathcal{C}_{noise}$ and each conformation $\mathcal{C}_{ref}^k$ in the reference conformation ensemble $\{C_{ref}^k\}_{k\in[1,\cdots,K]}$, where $K$ denotes the ensemble size. The reference conformation with the lowest RMSD is selected and this conformation is then used to compute the loss for backpropagation:

\begin{equation}
    k^* = {{\arg\min_{k \in \{{1, \dots, K\}}}}}~\text{RMSD}(\mathcal{C}_{noise},~\mathcal{C}_{ref}^k)
\end{equation}

\begin{equation}
    \text{Loss}=\text{LossFn}(\mathcal{C}_{gen},\mathcal{C}_{ref}^{k^*})
\end{equation}

\subsection{Objective functions}
Our objective function is defined as the sum of the permutation-invariant root mean square deviation (piRMSD) and internal coordinates (IC) loss, as expressed in the following equation:
\begin{equation}
\text{LossFn}=\mathcal{L}_{\text{piRMSD}}+\mathcal{L}_{ic}
\end{equation}

\subsubsection{Permutation-invariant RMSD loss} The Root Mean Square Deviation (RMSD) is commonly used as the evaluation metric for assessing molecular conformations. RMSDs are typically calculated rigidly, following the atomic position order as listed in coordinate files or data structures. Mathematically, Suppose ${\mathcal{C}_{ref}}=\left\{ c_{1},c_{2},\cdots ,c_{N} \right\}$ is a ground truth conformation and ${\mathcal{C}_{gen}}=\left\{ c'_{1},c'_{2},\cdots ,c'_{N} \right\}$ is an aligned generated conformation, the standard RMSD formula of can be defined as:
\begin{equation}
\text{RMSD}(\mathcal{C}_{ref}, \mathcal{C}_{gen}) = \sqrt{\frac{1}{N}\sum_{i=1}^{N}\left\| c_{i}-c'_{i} \right\|^2}
\end{equation}

\begin{figure}[!b]
    \centering
    \includegraphics[width=0.46\textwidth]{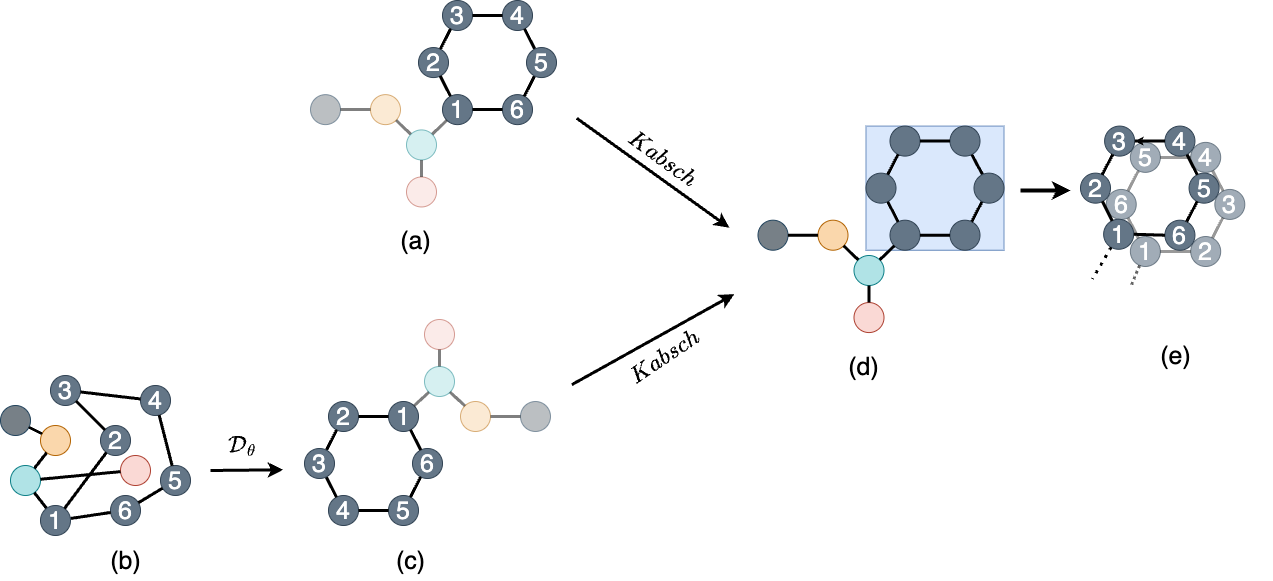}
    \caption{\textbf{An example of substructure symmetries in molecular conformation generation} Atom types are distinguished by different colors, while bond types are depicted uniformly without differentiation. (a) represents the ground truth conformation, (b) is the initial noise, (c) is generated from (b) via our model, (d) and (e) show the results after alignment, with (e) highlighting alignment issues caused by substructure symmetries.}
    \label{fig3}
\end{figure}

Before calculating RMSD, it is essential to align molecules to eliminate differences due to translation and rotation. However, the alignment process, typically performed using the Kabsch algorithm \parencite{lawrence2019purely}, often overlooks substructure symmetries, as illustrated in Figure \ref{fig3}. This oversight can result in predicted substructures being compared to incorrect counterparts, leading to inaccuracies in RMSD calculations. To address this issue, we propose piRMSD as our training loss function.

The piRMSD determines the minimum RMSD by identifying the optimal permutation of symmetrical substructures within conformations. The loss function is defined as follows:
\begin{equation}
\mathcal{L}_\text{piRMSD}(\mathcal{C}_{ref},\mathcal{C}_{gen})=\underset{\mathcal{T} \in S}{\min} ~\text{RMSD}(\mathcal{C}_{ref}, \mathcal{T}*\mathcal{C}_{gen})
\end{equation}
where $S$ denotes the set of symmetric permutation schemes extracted from the molecular graph structure:
\begin{equation}
S = \left\{ \mathcal{T}_1:\left\{ (c_2,c_3),(c_5,c_6) \right\},\cdots, \mathcal{T}_n:\left\{\cdots\right\}\right\}
\end{equation}
where $\mathcal{T}_{i}$ represents a scheme composed of all permutable atom mappings within a symmetric substructure, such as atom swapping of $(c_2,c_3)$ and $(c_5,c_6)$. The piRMSD is defined as the minimum RMSD obtained after implementing all such schemes. For further details, refer to the Appendix. 

\begin{table*}[h]
\centering
\begin{tabular}{llcccccccc}
\toprule
\multirow{2}{*}{Method} & & \multicolumn{4}{c}{Recall} & \multicolumn{4}{c}{Precision} \\
\cmidrule(lr){3-6} \cmidrule(lr){7-10}
& & \multicolumn{2}{c}{Coverage $\uparrow$} & \multicolumn{2}{c}{AMR $\downarrow$} & \multicolumn{2}{c}{Coverage $\uparrow$} & \multicolumn{2}{c}{AMR $\downarrow$} \\
& & Mean & Med & Mean & Med & Mean & Med & Mean & Med \\
\midrule
RDKit & & 85.1 & \textbf{100.0} & 0.235 & 0.199 & 86.8 & \textbf{100.0} & 0.232 & 0.205 \\
OMEGA & & 85.5 & \textbf{100.0} & \underline{0.177} & \underline{0.126} & 82.9 & \textbf{100.0} & 0.224 & 0.186 \\
GeoMol & & 91.5 & \textbf{100.0} & 0.225 & 0.193 & 86.7 & \textbf{100.0} & 0.270 & 0.241 \\
GeoDiff & & 76.5 & \textbf{100.0} & 0.297 & 0.229 & 50.0 & \underline{33.5} & 0.524 & 0.510 \\
Torsional diffusion & & \underline{92.8} & \textbf{100.0} & 0.178 & 0.147 & \underline{92.7} & \textbf{100.0} & 0.221 & 0.195 \\
\midrule
EDM & & 60.8 & \underline{60.0} & 0.374 & 0.321 & 80.5 & \textbf{100.0} & \underline{0.220} & \textbf{0.040}\\
EquiBoost & & \textbf{93.2} & \textbf{100.0} & \textbf{0.154} & \textbf{0.092} & \textbf{94.5} & \textbf{100.0} & \textbf{0.141} & \underline{0.081} \\
\bottomrule
\end{tabular}
\caption{Performance of conformation generation methods on the GEOM-QM9 dataset, evaluated by Coverage (\%) and Average Minimum RMSD (\r{A}). The Coverage threshold is set to $\delta=0.5$ \r{A}. We use \textbf{bold} to highlight the optimal results and \underline{underline} to denote the second-best results.}
\label{tab1}
\end{table*}

\begin{table*}[ht]
\centering
\begin{tabular}{llcccccccc}
\toprule
\multirow{2}{*}{Method} & & \multicolumn{4}{c}{Recall} & \multicolumn{4}{c}{Precision} \\
\cmidrule(lr){3-6} \cmidrule(lr){7-10}
& & \multicolumn{2}{c}{Coverage $\uparrow$} & \multicolumn{2}{c}{AMR $\downarrow$} & \multicolumn{2}{c}{Coverage $\uparrow$} & \multicolumn{2}{c}{AMR $\downarrow$} \\
& & Mean & Med & Mean & Med & Mean & Med & Mean & Med \\
\midrule
RDKit ETKDG & & 38.4 & 28.6 & 1.058 & 1.002 & 40.9 & 30.8 & 0.995 & 0.895 \\
OMEGA & & 53.4 & 54.6 & 0.841 & 0.762 & 40.5 & 33.3 & 0.946 & 0.854 \\
GeoMol & & 44.6 & 41.4 & 0.875 & 0.834 & 43.0 & 36.4 & 0.928 & 0.841 \\
GeoDiff & & 42.1 & 37.8 & 0.835 & 0.809 & 24.9 & 14.5 & 1.136 & 1.090 \\
Torsional Diffusion & & \textbf{72.7} & \textbf{80.0} & \textbf{0.582} & \textbf{0.565} & \underline{55.2} & \underline{56.9} & \underline{0.778} & \underline{0.729} \\
\midrule
EquiBoost & & \underline{66.9} & \underline{72.5} & \underline{0.645} & \underline{0.627} & \textbf{60.9} & \textbf{68.0} & \textbf{0.708} & \textbf{0.664} \\
\bottomrule
\end{tabular}
\caption{Performance of conformation generation methods on the GEOM-DRUGS dataset, evaluated by Coverage (\%) and Average Minimum RMSD (\r{A}). The Coverage threshold is set to $\delta=0.75$ \r{A}. We use \textbf{bold} to highlight the optimal results and \underline{underline} to denote the second-best results.}
\label{tab2}
\end{table*}

\subsubsection{Internal coordinate loss} In addition to Cartesian coordinates, internal coordinates (IC) are another representation of molecular geometry systems. Our IC Loss $\mathcal{L}_{ic}$ consists of four terms: bond length, bond angle, dihedral angle, and atom Euclidean distance. For detailed calculations of each term, refer to the Appendix.

By incorporating these two loss functions, our model not only directly optimizes the global coordinates of conformations but also constrains internal structures such as bond lengths and bond angles, facilitating the generation of more physically plausible molecular conformations.

\section{Experiment}
We evaluate EquiBoost by comparing the generated conformations and the ground truth. The first section details the experimental setup, including datasets, baseline methods, and evaluation metrics. The subsequent section provides a thorough analysis of the experimental results. Additional results, including ablation studies, are provided in Appendix.
\subsection{Experiment setup} 
\subsubsection{Dataset}
Following previous work, We evaluate our method on the GEOM-QM9 \parencite{ramakrishnan2014quantum} and GEOM-DRUGS \parencite{axelrod2022geom} datasets. The GEOM-QM9 dataset includes 133,258 molecules, each with up to 9 heavy atoms. For GEOM-DRUGS, the dataset comprises mid-sized drug-like molecules with a maximum of 91 heavy atoms per molecule, encompassing 304,466 molecules. We adopt the train/validation/test split from \parencite{ganea2021geomol}, with 106,586/13,323/1,000 molecules for GEOM-QM9 and 243,473/30,433/1,000 molecules for GEOM-DRUGS.

\subsubsection{Baselines}
 We compare our method with the most recent and robust existing methods. RDKit/ETKDG \parencite{riniker2015better}, one of the most widely used open-source packages for molecular conformation generation, is included in our evaluation. We also consider OMEGA \parencite{hawkins2012conformer}, a commercial tool known for its efficacy in this domain. In addition to these cheminformatics approaches, we also evaluate machine learning-based methods like GeoDiff \parencite{xu2022geodiff}, GeoMol \parencite{ganea2021geomol}, and Torsional diffusion \parencite{jing2022torsional}. The performance data for these models are sourced from the Torsional Diffusion paper. Furthermore, as part of our baseline, we apply the generative diffusion framework known as EDM \parencite{karras2022elucidating} to the task of molecular conformation generation for the first time.

\begin{table}[hb]
\centering
\begin{tabular}{llcccccccc}
\toprule
Methods & Steps & AMR-R & AMR-P & Runtime \\
\midrule
RDKit ETKDG & - & 1.002 & 0.895 & 0.1 \\
GeoDiff & 5000 & 0.809 & 1.090 & 305 \\
\midrule
    \multirow{3}{*}{Torsional Diffusion} & 20 & \textbf{0.565} & 0.729 & 4.90 \\
                            & 10 & 0.580 & 0.791 & 2.82 \\
                            & \textbf{5} & 0.685 & 0.963 & 1.76 \\
\midrule
EquiBoost & \textbf{5} & 0.627 & \textbf{0.664} & \textbf{1.17}\\
\bottomrule
\end{tabular}
\caption{Median AMR and runtime(core-secs per conformer) of existing methods, evaluated on CPU for comparison with RDKit.}
\label{tab3}
\end{table}

\subsubsection{Evaluation metrics}
In this task domain, the quality and diversity of conformation generation are crucial evaluation aspects. \parencite{ganea2021geomol} employed four types of metrics to evaluate conformation ensembles, all based on root-mean-square deviation(RMSD) between generated and ground-truth atomic coordinates. Let $\{\mathcal{C}_k\}_{k\in[1..K]}$ and $\{\mathcal{C}_t\}_{t\in[1..T]}$ denote the set of generated conformations and ground-truth conformations, respectively. In our approach, $K$ is set to be twice of $T$. The metrics are defined as follows:
\begin{align}
\text{COV-R}=&\frac{1}{T} | \{ t \in [1 .. T] : \exists k \in [1 .. K],\notag \\
&~~~~~~~~~~\text{RMSD}(\mathcal{C}_k, \mathcal{C}_t) < \delta \} |\\
\text{AMR-R}=&
\frac{1}{T}\sum_{t\in[1..T]}\min_{k\in[1..K]} \text{RMSD}(\mathcal{C}_k,\mathcal{C}_t)
\end{align}
where $\text{COV}$ stands for coverage, $\text{AMR}$ stands for Average Minimum RMSD. The suffix R represents Recall, which measures how effectively the model identifies actual positive cases (ground truth conformations). The coverage threshold $\delta$ determines whether a prediction is considered correct, set to 0.5 \r{A} and 0.75 \r{A} respectively for GEOM-QM9 and GEOM-DRUGS datasets. The other two metrics, $\text{COV-P}$ and $\text{AMR-P}$, are similarly defined but swap the set of generated conformations and ground-truth conformations. Here, the suffix P represents precision, which assesses the quality of the generated conformations by evaluating their fidelity to the ground-truth conformations.

\subsection{Results and discussion}
EquiBoost outperforms all previously established methods on the GEOM-QM9 dataset (Table \ref{tab1}). Compared to the previous state-of-the-art diffusion method, EquiBoost notably reduces AMR-P Mean and AMR-P Median by 36.2\% and 58.5\%, respectively. On the GEOM-DRUGS dataset (Table \ref{tab2}), EquiBoost surpasses all prior methods in Precision, achieving an impressive COV-P with a mean of 60.9\% and a median of 68\%. In terms of Recall, it ranks second only to Torsional Diffusion, surpassing all traditional methods and most contemporary machine learning models. EquiBoost demonstrates exceptional accuracy and diversity in conformation generation, indicating its potential for application in computational drug design.

Furthermore, we apply the EDM framework to our task. The results indicate that EDM achieved superior AMR-P performance on the GEOM-QM9 dataset compared to previous work. However, EquiBoost remains superior overall, highlighting its potential to supplant diffusion models in certain generative tasks.

As a boosting model, EquiBoost exhibits substantial sampling efficiency (Table \ref{tab3}). It achieves superior accuracy in just 5 sampling steps compared to GeoDiff with 5000 steps and Torsional Diffusion 20 steps. This efficiency enables EquiBoost to produce higher-quality molecular conformations in less time, making it highly advantageous for practical applications.

Torsional Diffusion begins with RDKit-generated initial conformations, effectively capturing conformation diversity. Similarly, EquiBoost leverages RDKit for conformation initialization, thereby naturally inheriting this advantage. Although EquiBoost achieves the highest accuracy without relying on RDKit, this comes at the cost of a substantial reduction in diversity (see Appendix for details). Given that the GEOM dataset is derived from RDKit-initialized conformations refined with semi-empirical density functional theory (DFT), the conformation diversity observed in this dataset might be “pseudo”, raising concerns about the reliability of recall assessments based on it.

Our model operates directly in Euclidean space, optimizing local structures while utilizing RDKit-initialized conformations. Torsional Diffusion, on the other hand, optimizes torsional angles while keeping local structures fixed. This distinction likely accounts for the superior precision of our model compared to Torsional Diffusion.

\section{Conclusion}
We propose EquiBoost, an equivariant boosting model designed for molecular conformation generation. EquiBoost outperforms traditional cheminformatics and machine learning methods on the GEOM dataset, achieving not only higher accuracy but also greater sampling efficiency. Notably, by employing a boosting model for molecular conformation generation, we achieve results superior to those of diffusion models, thus revitalizing the boosting approach and revealing its potential as a powerful alternative to diffusion models in certain scenarios. 

\subsubsection{Future work} The model's capabilities should be further validated through implementation in real-world applications. For instance, applying EquiBoost to molecular docking could enhance accuracy or improve access rates in virtual screening. Given EquiBoost's competitive performance in conformation generation, extending its application to larger and more complex molecular systems is a promising direction. Beyond molecular systems, the principles underlying EquiBoost could be applied to various other fields where generative models are crucial. These include, but are not limited to, image generation, music composition, and other domains where the balance between generation quality and computational efficiency is vital.

\printbibliography[heading=subbibliography, title={References}, segment=0]

\newrefsegment

\newpage
\appendix
\section{Details of Method}
\subsection{Equiformer in molecular conformation generation}
Equiformer \parencite{liao2023equiformerv2} serves as the foundational equivariant model in our algorithm. It consists of multiple layers, including layer normalization, equivariant graph attention, and feed-forward network. Below, we describe how this model ensure equivariance and be applied in molecular conformation generation.

We start by defining a special type of feature known as the irreducible representation (irreps). The irreps feature $f$ is a special type of feature that can be decomposed into scalar features $f^{\text{scalar}}$ and vector features $f^{\text{vector}}$. Under SE(3) transformations, the scalar features are invariant while the vector features are equivariant.

Next, we explain how to generate irreps features from a graph $\mathcal{G}$ and its conformation $\mathcal{C}$. Given a source atom $a_i$ and a target atom $a_j$, the initial message $m_{ij}$ in the graph neural network is derived from the source atom embedding $x_i$, the target atom embedding $x_j$ and the edge embedding $e_{ij}$. These embeddings are based on atom type and bond type, making them scalar values that are invariant to SE(3) transformations. To incorporate geometric information, $m_{ij}$ is combined with the relative positions to obtain irreps features $f_{ij}$:
\begin{equation}
    f_{ij}=m_{ij}\otimes_{\omega(||{\overset{\rightarrow}{r_{ij}}}||)}^\text{DTP}~\text{SH}({\overset{\rightarrow}{r_{ij}}}) \\
\end{equation}
Here, DTP refers to the depth-wise tensor product. The relative position $\overset{\rightarrow}{r_{ij}}=c_i-c_j$ is first embedded by spherical harmonics ($\text{SH}$) and then interacts with the initial message through DTP. The interaction is weighted by $\omega$, which is parameterized by the norm of the relative position vector $||\overset{\rightarrow}{r_{ij}}||$. For the detailed explanation of underlying principles, refer to the Equiformer paper \parencite{liao2023equiformer}.

Give the equivariant feature $f_{ij}$ as input, a Multi-Layer Perceptron Attention (MLPA) is applied to compute the attention score $a_{ij}$ and values $v_{ij}$. Since attention weights $a_{ij}$ determine the interaction between each node and its neighboring nodes, they must be invariant under SE(3) transformations. Therefore, only scalar features $f_{ij}^{\text{scalar}}$ are used to calculate the attention weights:

\begin{align}
    z_{ij} &= a^\top\text{LeakyReLU}(f_{ij}^{\text{scalar}}) \\
    a_{ij} &= \text{softmax}_j(z_{ij})=\frac{\text{exp}(z_{ij})}{\sum_{\hat{j}\in\mathcal{N}(i)}\text{exp}(z_{i\hat{j}})}
\end{align}
where $a$ is a learnable vector, and $z_{ij}$ is a scalar obtained by taking the dot product between $a$ and the LeakyReLU-activated scalar features $f_{ij}^{\text{scalar}}$. The attention weights $a_{ij}$ are then derived using a softmax function applied over all neighboring nodes $j\in\mathcal{N}(i)$.

The values $v_{ij}$ represent features transmitted between nodes, and they are equivariant with respect to the input features that undergo transformations. These values can be computed as follows:

\begin{align}
\mu_{ij} &= \text{Gate}(f_{ij}) \\
v_{ij} &= \text{Linear}([\mu_{ij} \otimes_{\omega}^{\text{DTP}} \text{SH}(\overset{\rightarrow}{r_{ij}})])
\end{align}
The Gate function introduces non-linearity, while the Linear function acts as a multi-layer perceptron tailored for irreps features. 

After applying attention to obtain both the attention score $a_{ij}$ and values $v_{ij}$, the updated messages $m_{ij}'$ are computed as:
\begin{equation}
    m_{ij}'=a_{ij} \cdot v_{ij}
\end{equation}
Since $a_{ij}$ is invariant and $v_{ij}$ is equivariant, their product, $m_{ij}'$, remains equivariant. The updated features $x_i'$ of atom $a_i$ can be obtained by:

\begin{equation}
    x_i'=\sum_{j\in\mathcal{N}(i)}m_{ij}'
\end{equation}
The summation operation over $m_{ij}'$ does not affect its equivariance, so $x_o'$ continues to be equivariant. We denote this process of equivariant graph attention (EGA) at layer $l$ as $x^{l}=\text{EGA}(x^{l-1})$. For an equivariant graph transformer with $L$ layers, The vector component, $x^{L, \text{vector}}$, contains different types of equivariant features, with the displacement of atom coordinates $\Delta{\mathcal{C}}$ being a type-1 equivariant feature. From the outputs of the model $\gamma_\theta$, we extract $\Delta{\mathcal{C}}$ from $x^{L, \text{vector}}$, described as follows:

\begin{align}
x^{L, \text{scalar}}, x^{L, \text{vector}} &= \gamma_\theta(\mathcal{G}, \mathcal{C}_{\text{noise}}) \\
\Delta \mathcal{C} &= \text{extract}_{\Delta \mathcal{C}}(x^{L, \text{vector}})
\end{align}

Finally, the generated conformation $\hat{\mathcal{C}}$ is obtained by adding $\Delta{\mathcal{C}}$ to the initial noisy conformation $C_{noise}$.

\subsection{EquiBoost framework equivariance}
Generating $\hat{\mathcal{C}}$ from $\mathcal{C}_{noise}$ using a neural network $\gamma_\theta$ requires the model to be equivariant. This requirement remains intact when applying the boosting mechanism. Below, we prove that EquiBoost maintains equivariance if each weak learner $\gamma_\theta$ is equivariant. EquiBoost can be illustrated by the following equation:
\begin{equation}
    \mathcal{D}(\mathcal{C}|\mathcal{G},\mathcal{C}_{noise})=\mathcal{C}_{noise}+\sum_{m=0}^{M-1}{\gamma_{\theta,m}(\mathcal{G},\mathcal{C}^m)},
\end{equation}
if we apply any SE(3) transformation $g$ on the input noisy conformation $\mathcal{C}_{noise}$, then we have:
\begin{align}
\mathcal{D}(\mathcal{C}|\mathcal{G},g\cdot\mathcal{C}_{noise})&=g\cdot\mathcal{C}_{noise}+\sum_{m=0}^{M-1}{\gamma_{\theta,m}(\mathcal{G},g\cdot\mathcal{C}^m)} \notag \\
&=g\cdot\mathcal{C}_{noise}+g\cdot\sum_{m=0}^{M-1}{\gamma_{\theta,m}(\mathcal{G},\mathcal{C}^m)} \notag \\
&= g\cdot(\mathcal{C}_{noise}+\sum_{m=0}^{M-1}{\gamma_{\theta,m}(\mathcal{G},\mathcal{C}^m)}) \notag\\
&=g\cdot\mathcal{D}(\mathcal{C}|\mathcal{G},\mathcal{C}_{noise})
\end{align}
Therefore, our EquiBoost model $\mathcal{D}(\mathcal{C}|\mathcal{G},\mathcal{C}_{noise})$ is also equivariant.

\subsection{Higher-order adjacency matrix}
\begin{figure}[!h]
    \centering\includegraphics[width=0.45\textwidth]
    {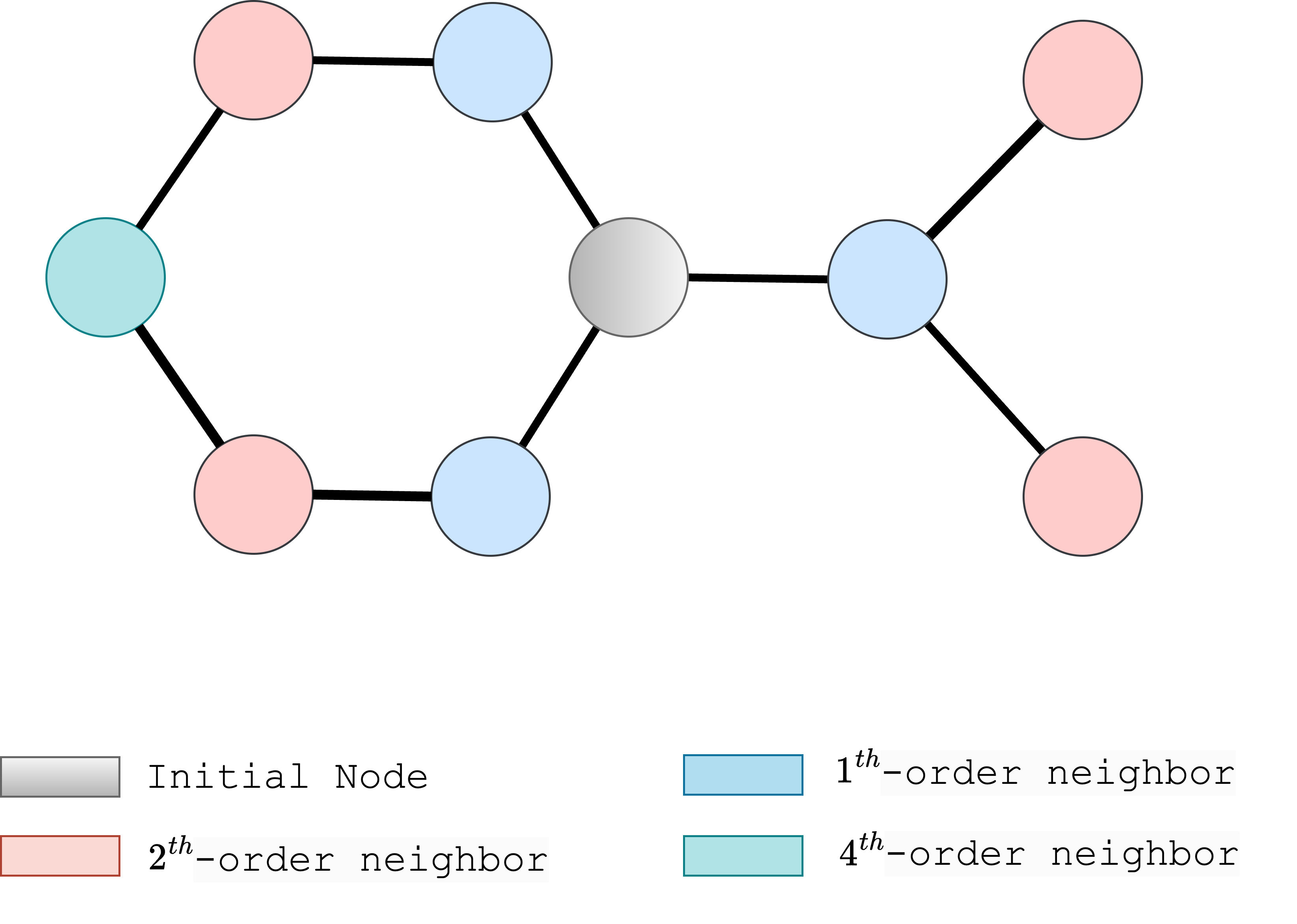}
    \caption{\textbf{Example of Higher-order neighbors.}}
    \label{figS1}
\end{figure}

We utilize an adjacency matrix to represent the existence of chemical bonds between atoms in a molecule. To capture more extensive connectivity information, we extend the adjacency matrix to include higher-order adjacency relationships, as shown in Figure \ref{figS1}. Specifically, we incorporate $h^{th}$-order neighbors as edge attributes in the molecular graph $(h \in \mathbb{Z}^+)$, allowing the model to extract latent information from these higher-order neighbors. Theoretically, the inclusion of these higher-order adjacency relationships facilitates faster convergence during training, thereby improving efficiency and robustness.

\subsection{Loss function}
\subsubsection{Permutation-invariant RMSD}

\begin{figure*}[t]
    \centering
    \includegraphics[width=0.9\textwidth]{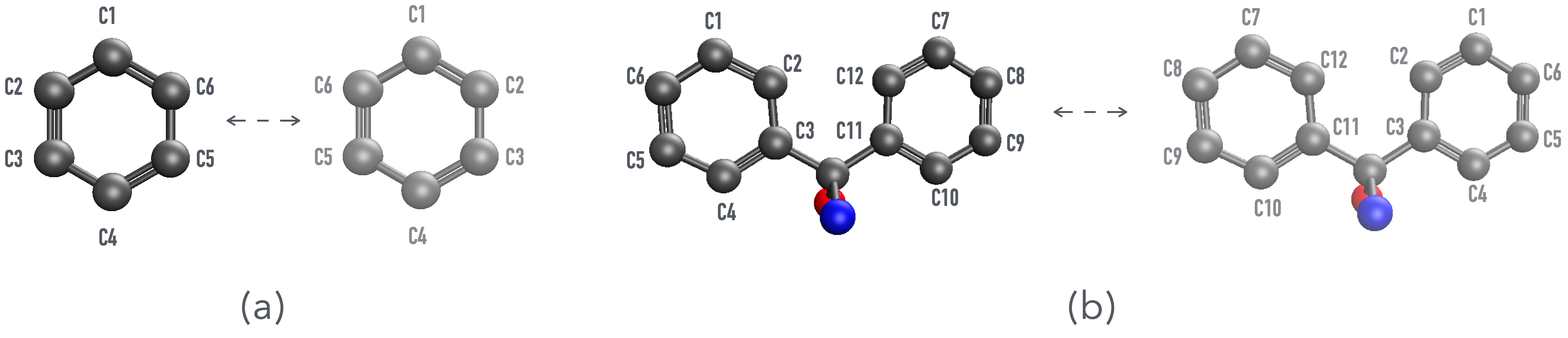}
    \caption{\textbf{Example of Node Symmetry in Molecular Conformations} (a) A self-symmetric ring structure. (b) A symmetric substructure connected to the same node.}
    \label{figS2}
\end{figure*}

Before calculating Permutation-Invariant RMSD, We need to identify the ensemble of symmetric substructures $S$ in the molecular graph. Specifically, we use a breadth-first search to identify self-symmetric ring structures, as shown in Figure \ref{figS2}(a), and depth-first search to find symmetric substructures connected to the same node, as shown in Figure \ref{figS2}(b). Following this, the optimal permutation scheme $\mathcal{T}$ must be determined. Theoretically, we need to compute the RMSD for all permutation plans to select $\mathcal{T}$, which is computationally inefficient; hence, we adopt an approximation method as an alternative.

Algorithm 1 details the permutation-invariant RMSD loss. For the ground truth conformation $\mathcal{C}$ and generated conformation $\hat{\mathcal{C}}$, we first rearrange the ensemble $S$ so that each substructure $s$ is ordered in descending order based on the number of atoms. For each symmetric substructure ${s}$, we then calculate their centroids and determine the pairwise distances between corresponding centroids in $\mathcal{C}$ and $\hat{\mathcal{C}}$. The Hungarian algorithm is applied to the distance matrix to obtain the pairing that minimizes the sum of these centroids distances, and the order of atoms in $\hat{\mathcal{C}}$ is rearranged accordingly. Upon completing the permutation of all symmetric substructures, we calculate the standard RMSD between $\mathcal{C}$ and the rearranged conformation $\hat{\mathcal{C}}$ as the piRMSD.

\begin{algorithm}
\caption{Permutation-Invariant RMSD}
\begin{algorithmic}[1]
\Require Ground truth conformation $\mathcal{C}$, generated conformation $\hat{\mathcal{C}}$, an ensemble of symmetric substructures $S = \{s_1:\{\cdots\}, \dots, s_n:\{\cdots\}\}$
\Ensure Permutation-invariant RMSD $\mathcal{L}_{\text{piRMSD}}$

\State Rearrange the symmetric substructures {s} in {S} so that {s} follow a descending order of the number of atoms.
\For{$s \leftarrow 1$ \textbf{to} $s_{max}$}
    \State Calculate centroids of substructures in $s$
    \State Compute the distance matrix $\mathcal{M}$ between centroids
    \State Determine the optimal permutation using the Hungarian algorithm to minimize centroid distances
    \State Permute the substructures
\EndFor
\State Calculate RMSD between $\mathcal{C}$ and rearranged $\hat{\mathcal{C}}$

\end{algorithmic}
\end{algorithm}

\subsubsection{Internal coordinate loss}
To comprehensively describe molecule conformations, we integrate Internal Coordinate(IC) Loss into our loss function. We compute IC loss between predicted and target molecular conformations by considering the differences in bond lengths, bond angles, dihedral angles, and inter-atomic distances. The equation is defined as follows:
\begin{equation}
    \mathcal{L}_\text{ic} = \mathcal{L}_\text{bond~length} + \mathcal{L}_\text{bond~angle} + \mathcal{L}_\text{dihedral} + \mathcal{L}_\text{edist}
\end{equation}

\begin{equation}
\begin{aligned}
    \mathcal{L}_{\text{bond~length}} &= \text{SE}(B^{\text{length}}_{ref}, B^{\text{length}}_{gen}) \\
    \mathcal{L}_{\text{bond~angle}} &= \text{SE}(B^{\text{angle}}_{ref}, B^{\text{angle}}_{gen}) \\
    \mathcal{L}_\text{dihedral} &= \text{SE}(D^\text{angle}_{ref}, D^\text{angle}_{ref}) \\
    \mathcal{L}_\text{edist} &= \text{MSE}(E^\text{dist}_{ref},E^\text{dist}_{gen})
\end{aligned}
\end{equation}

The process begins by converting the atom adjacency matrix, which contains chemical bond information, into Z-matrix form. Using information from the Z-matrix, both the ground-truth and model-predicted Cartesian coordinates are converted into internal coordinates, including bond lengths, bond angles, and dihedral angles.
For inter-atomic distances, the Mean Squared Error (MSE) loss is calculated between the predicted and target distance matrices ($E^{\text{dist}}$). For bond lengths ($B^{\text{length}}$), bond angles ($B^{\text{angles}}$), and dihedral angles ($D^{\text{length}}$), the Squared Error (SE) loss is utilized. Specifically, for dihedral angles, if the Absolute Error(AE) exceeds $\pi$ (180 degrees), the difference is adjusted to $(2\pi- \text{AE})$ to ensure the angle difference remains within a reasonable range. 

\subsection{Details of training and inference}
In this section, we present pseudocode to illustrate the training and inference processes of EquiBoost and EDM.

\subsubsection{EquiBoost}

There are two methods for initializing conformations in EquiBoost: Random Sampling (RS) and Constrained Random Sampling (CRS). We use RS and CRS as the suffix of EquiBoost to distinguish them. Below we give the detailed algorithms of RS and CRS. Whether using CRS is set to be a hyperparameter and is not distinguished in the notation when calling the random sampling algorithm. The random sampling algorithm is shown in Algorithm \ref{rs}.

\begin{algorithm}
\caption{RandomSample}
\begin{algorithmic}[1]
\Require Molecular graph $\mathcal{G}$
\Ensure Initialized noisy conformation $\mathcal{C}_{noise}$
\If{using random sampling}
    \State $\mathcal{C}_{noise} \sim \mathcal{N}(0, \mathbf{I_{N\times3}})$
\Else
    \State Initialize $(\xi, \tau)$ of $\mathcal{C}_{noise}$ using RDKit
    \State Randomize the torsional angles $\tau$
\EndIf

\State\Return $\mathcal{C}_{noise}$
\end{algorithmic}
\label{rs}
\end{algorithm}

Using a single molecule as an example, we provide the pseudo code for training in Algorithm \ref{eb_train}. To enhance efficiency in practice, we utilize mini-batch training. Additionally, to implement the randomization of the boost steps, we sample an $M_{train}$ to replace the fixed boost steps $M$ during the training phase.

\begin{algorithm}
\caption{EquiBoost Training}
\begin{algorithmic}[1]
\Require Molecular graph $\mathcal{G}$, Ensemble of molecular conformations $\{C_{ref}^k\}_{k \in [1, \dots, K]}$, weak learners $\gamma_\theta$
\Ensure Loss value $\mathcal{L}$
\Statex \# $\mathcal{C}_0$ is identical to $\mathcal{C}_{noise}$
\State $\mathcal{C}_{noise}\leftarrow\text{RandomSample}(\mathcal{G})$
\Statex \# Using boosting to generate conformation $\mathcal{C}_{M}$
\State $M_{train} \leftarrow \text{uniform}(0, M-1)$
\For{$m \leftarrow 0$ \textbf{to} $M_{train}-1$}
    \State $\mathcal{C}^{m+1} \leftarrow \gamma_\theta(\mathcal{G}, \mathcal{C}^{m})$
\EndFor
\Statex \# Optimal conformation mapping
\For{$C_{ref}^k \in \{C_{ref}^k\}_{k \in [1, \dots, K]}$}
\State Compute RMSD$(\mathcal{C}_{noise}, C_{ref}^k)$
\EndFor
\State Select $k^* = \arg\min_k \text{RMSD}(\mathcal{C}_{noise}, C_{ref}^k)$
\Statex \# Loss calculation
\State $\mathcal{L} = \text{LossFn}(\mathcal{C}^{M}, C_{ref}^{k^*})$
\State\Return $\mathcal{L}$
\end{algorithmic}
\label{eb_train}
\end{algorithm}

The inference algorithm is detailed in Algorithm \ref{eb_infer}, where $M$ is fixed during the inference stage.

\begin{algorithm}
\caption{EquiBoost Inference}
\begin{algorithmic}[1]
\Require molecular graph $\mathcal{G}$, trained model $\mathcal{D}_\theta$
\Ensure generated conformation $\mathcal{C}_M$
\State $\mathcal{C}_{noise}\leftarrow\text{RandomSample}(\mathcal{G})$
\For{$m \leftarrow 0$ to $M-1$}
    \State $\mathcal{C}_{m+1} \leftarrow \gamma_\theta(\mathcal{G}, \mathcal{C}_m)$
\EndFor
\State \Return generated conformation $\mathcal{C}_M$
\end{algorithmic}
\label{eb_infer}
\end{algorithm}

\subsubsection{EDM}

We apply the EDM model \parencite{karras2022elucidating}, a diffusion-based generative framework, to molecular conformation generation for the first time. This approach serves as one of our baselines, providing comparisons to previous related work and our main model, EquiBoost. The EDM model begins with random noise as input and generates molecular conformations. Like all diffusion models, EDM consists of a forward and a reverse process. During the forward diffusion process, random noise $n\sim\mathcal{N}(0,\sigma^{2}I_{N\times3})$ with noise level $\sigma$ is gradually added to the Cartesian coordinates of the conformer $\mathcal{C}$, resulting in the noisy conformer $\mathcal{C}_0\sim \mathcal{N}(0,\sigma_{max}^2I_{N\times3})$. In the reverse diffusion process, we train a model $f_{\theta}(x;\sigma)$ that predicts the coordinate offset relative to the ground truth conformer $\mathcal{C}$. The model $f_{\theta}$ functions as a denoising model, where $\theta$ represents the model parameters. The detailed pseudocode for EDM training and sampling is provided in Algorithms 5 and 6, respectively.

\begin{algorithm}
\caption{EDM Training}
\begin{algorithmic}[1]
\Require Molecular conformations $\{C_{ref}^k\}_{k\in[1,\cdots,K]}$, Distribution of training noise levels $p(\sigma)$, weighting function $\lambda(\sigma)$
\Ensure Loss value $\mathcal{L}$

\State Sample a conformation $\{\mathcal{C}\}$ from  $\{C_{ref}^k\}_{k\in[1,\cdots,K]}$
\State Sample standard deviations $\sigma$ from $p(\sigma)$
\State Sample noise vectors $\mathbf{n} \sim \mathcal{N}(0, \sigma^2 \mathbf{I}_{N\times3})$
\State Get perturbed data $\tilde{\mathcal{C}} = \mathcal{C} + \mathbf{n}$
\Statex \# Loss calculation
\State $\mathcal{L} = \text{LossFn}(\mathcal{C}, \tilde{\mathcal{C}}+f_{\theta}(\tilde{\mathcal{C}},\sigma))$
\State \Return $\mathcal{L}$
\end{algorithmic}
\end{algorithm}

\begin{algorithm}
\caption{EDM Sampling (Heun's 2nd Order Method)}
\begin{algorithmic}[1]
\Require Maximum variance $\sigma_{\max}^2$, Number of timesteps $T$, trained model $f_\theta$
\Ensure Final sampled conformation $\hat{\mathcal{C}}$

\State $\mathcal{C}_{0} \sim \mathcal{N}(0, \sigma_{\max}^2 \mathbf{I}_{N\times3})$
\For {$i = 0, \ldots, T-1$}
    \State $\Delta{\mathcal{C}}_i = (\mathcal{C}_i - \mathbf{f}_\theta(\mathcal{C}_i, t_i)) / t_i$
    \State $\mathcal{C}_{i+1} = \mathcal{C}_i + (t_{i+1} - t_i) \Delta{\mathcal{C}}_i$
    \If {$t_{i+1} > 0$}
        \State $\Delta{\mathcal{C}}_i' = (\mathcal{C}_{i+1} - \mathbf{f}_\theta(\mathcal{C}_{i+1}, t_{i+1})) / t_{i+1}$
        \State $\mathcal{C}_{i+1} = \mathbf{x}_i + (t_{i+1} - t_i) (\frac{1}{2} \Delta{\mathcal{C}}_i + \frac{1}{2} \Delta{\mathcal{C}}_i')$
    \EndIf
\EndFor
\end{algorithmic}
\end{algorithm}

\section{Details of experiment}

\begin{table*}[h]
\centering
\begin{tabular}{llcccccccc}
\toprule
\multirow{2}{*}{Method} & & \multicolumn{4}{c}{Recall} & \multicolumn{4}{c}{Precision} \\
\cmidrule(lr){3-6} \cmidrule(lr){7-10}
& & \multicolumn{2}{c}{Coverage $\uparrow$} & \multicolumn{2}{c}{AMR $\downarrow$} & \multicolumn{2}{c}{Coverage $\uparrow$} & \multicolumn{2}{c}{AMR $\downarrow$} \\
& & Mean & Med & Mean & Med & Mean & Med & Mean & Med \\
\midrule
RDKit ETKDG & & 85.1 & \textbf{100.0} & 0.235 & 0.199 & 86.8 & \textbf{100.0} & 0.232 & 0.205 \\
OMEGA & & 85.5 & \textbf{100.0} & \underline{0.177} & \underline{0.126} & 82.9 & \textbf{100.0} & 0.224 & 0.186 \\
\midrule
GeoMol & & 91.5 & \textbf{100.0} & 0.225 & 0.193 & 86.7 & \textbf{100.0} & 0.270 & 0.241 \\
GeoDiff & & 76.5 & \textbf{100.0} & 0.297 & 0.229 & 50.0 & \underline{33.5} & 0.524 & 0.510 \\
EquiBoost-RS & & 71.4 & \underline{95.6} & 0.321 & 0.262 & 92.5 & \textbf{100.0} & \textbf{0.140} & \textbf{0.053} \\
\midrule
Torsional diffusion & & \underline{92.8} & \textbf{100.0} & 0.178 & 0.147 & \underline{92.7} & \textbf{100.0} & 0.221 & 0.195 \\
EquiBoost-CRS & & \textbf{93.2} & \textbf{100.0} & \textbf{0.154} & \textbf{0.092} & \textbf{94.5} & \textbf{100.0} & \underline{0.141} & \underline{0.081} \\
\bottomrule
\end{tabular}
\caption{Performance of conformation generation methods on the GEOM-QM9 dataset, evaluated by Coverage (\%) and Average Minimum RMSD (\r{A}). The Coverage threshold is set to $\delta=0.5$ \r{A}. We use \textbf{bold} to highlight the optimal results and \underline{underline} to denote the second-best results.}
\label{tab2}
\end{table*}

\begin{table*}[h]
\centering
\begin{tabular}{llcccccccc}
\toprule
\multirow{2}{*}{Method} & & \multicolumn{4}{c}{Recall} & \multicolumn{4}{c}{Precision} \\
\cmidrule(lr){3-6} \cmidrule(lr){7-10}
& & \multicolumn{2}{c}{Coverage $\uparrow$} & \multicolumn{2}{c}{AMR $\downarrow$} & \multicolumn{2}{c}{Coverage $\uparrow$} & \multicolumn{2}{c}{AMR $\downarrow$} \\
& & Mean & Med & Mean & Med & Mean & Med & Mean & Med \\
\midrule
RDKit ETKDG & & 38.4 & 28.6 & 1.058 & 1.002 & 40.9 & 30.8 & 0.995 & 0.895 \\
OMEGA & & 53.4 & 54.6 & 0.841 & 0.762 & 40.5 & 33.3 & 0.946 & 0.854 \\
\midrule
GeoMol & & 44.6 & 41.4 & 0.875 & 0.834 & 43.0 & 36.4 & 0.928 & 0.841 \\
GeoDiff & & 42.1 & 37.8 & 0.835 & 0.809 & 24.9 & 14.5 & 1.136 & 1.090 \\
EquiBoost-RS & & {34.2} & {25.8} & {1.134} & {1.123} & \textbf{71.6} & \textbf{94.7} & \textbf{0.624} & \textbf{0.554} \\
\midrule
Torsional Diffusion & & \textbf{72.7} & \textbf{80.0} & \textbf{0.582} & \textbf{0.565} & {55.2} & {56.9} & {0.778} & {0.729} \\
EquiBoost-CRS & & \underline{66.9} & \underline{72.5} & \underline{0.645} & \underline{0.627} & \underline{60.9} & \underline{68.0} & \underline{0.708} & \underline{0.664} \\
\bottomrule
\end{tabular}
\caption{Performance of conformation generation methods on the GEOM-DRUGS dataset, evaluated by Coverage (\%) and Average Minimum RMSD (\r{A}). The Coverage threshold is set to $\delta=0.75$ \r{A}. We use \textbf{bold} to highlight the optimal results and \underline{underline} to denote the second-best results.}
\label{tab3}
\end{table*}

\subsection{Data preprocessing}
To ensure a fair comparison with previous work, we adopt the same dataset, GEOM, which includes the GEOM-QM9 and GEOM-DRUGS datasets. For dataset splitting during model training, we follow the partitions established by Geomol \parencite{ganea2021geomol} and Torsional Diffusion \parencite{jing2022torsional}. It's important to note that GeoDiff \parencite{xu2022geodiff} used a custom subset of the GEOM dataset, preventing us from directly using its reported performance data. Instead, we utilize the standardized results provided by the Torsional Diffusion paper for our baseline models, avoiding the need for independent evaluations.

Our preprocessing pipeline includes several key steps:

\textbf{1. Hydrogen removal}: Consistent with previous work, Hydrogen atoms are removed from the molecular structures to focus on the heavy atoms. This operation reduces the computational burden on the model.

\textbf{2. Graph construction}: In the GEOM dataset, molecular information is provided in .pickle files. From the RDKit Mol object stored in these files, we extract atom types, bond types, and bonding relationships, which are crucial for constructing the molecular graph. Based on the extracted molecular information, we create bidirectional graphs representing the molecular conformations. 

\textbf{3. Conformation initialization}: We employ two methods to initialize molecular conformations:

\begin{itemize}
\item \textbf{Random Sampling}: Atom coordinates are randomly sampled from a Gaussian distribution $\mathcal{N}(0, \mathbf{I_N})$, where the subscript $N$ denotes the number of atoms in the molecule.

\item \textbf{Constrained randomization based on RDKit-generated conformations}: Conformations are initialized using RDKit, utilizing the RDKit Mol object as input. A constrained randomization is then applied, specifically targeting the molecular torsional angles.
\end{itemize}

These initialized conformations, along with the constructed molecular graph, serve as inputs for EquiBoost. We compare these two conformation initialization strategies in the Ablation study section.

\subsection{Training configurations}
For molecular conformation generation on the GEOM-DRUGS dataset, EquiBoost is trained for 120 epochs using 6 NVIDIA A100-SXM4-80GB GPUs, taking a total of 42 hours. The maximum memory usage per GPU is less than 11GB, with an average GPU utilization of approximately 47\%. For the GEOM-QM9 dataset, training is conducted for 100 epochs using 2 NVIDIA A100-SXM4-80GB GPUs, taking a total of 28 hours. The maximum memory usage is less than 5GB, with an average GPU utilization of approximately 27\%. The hyperparameters used during training are as follows: optimizer (Adam), initial learning rate (0.0002), weight decay (0.001), learning rate scheduler (cosine annealing), Number of iterations until the first restart (1000), higher-order adjacency matrix (3-order), boosting number (5), model parameters(2.2M).

\subsection{Ablation study}
We conduct ablation studies on certain mechanisms within EquiBoost. To improve experimental efficiency, the study on boost steps is carried out on the GEOM-QM9 dataset, which contains relative fewer and simpler molecules.

\subsubsection{Boost steps}
We evaluate the impact of varying boost steps on the performance of EquiBoost. When the boost steps is set to 0, it indicates the use of only a single weak learner without boosting. As shown in Table \ref{tab1}, generation diversity, measured by AMR-R, improves with increasing boost steps. In contrast, generation quality, measured by AMR-P, does not exhibit a linear relationship with boost steps; it increases up to $M=3$ and then declines. Overall, increasing the boost steps enhances the model performance. To balance computational efficiency with these model performance metrics, we select $M=5$ as the optimal number of boost steps for EquiBoost.

\begin{table}[h]
\centering
\begin{tabular}{ccc}
\toprule
\textbf{Boost steps} & \textbf{AMR-R} & \textbf{AMR-P} \\
\midrule
0   & 0.325 & 0.161 \\
1   & 0.312 & 0.125 \\
3   & 0.283 & \textbf{0.054} \\
5   & 0.279 & 0.074 \\
10  & \textbf{0.277} & 0.071 \\
\bottomrule
\end{tabular}
\caption{Median AMR values across different numbers of weak learner in EquiBoost. \textbf{Bold} indicates the optimal results.}
\label{tab1}
\end{table}

\subsubsection{Conformation initialization}
As introduced in Data preprocessing section, we employ two ways to initialize molecular conformations. Here, we assess the impact of these two approaches on model performance. To facilitate a clearer comparison, we divided the existing methods in the table with horizontal lines. From top to bottom, the categories are cheminformatics approaches, machine learning methods based on random sampling initialized conformations and those using constrained random sampling initialized conformations. On the GEOM-QM9 dataset, as shown in Table \ref{tab2}, EquiBoost-CRS (constrained randomization based on RDKit-generated conformations) demonstrates a clear advantage over the random sampling approach, EquiBoost-RS, in terms of generation diversity. Despite this, EquiBoost-CRS achieves state-of-the-art performance in terms of generation quality. A similar trend is observed on the GEOM-DRUGS dataset, as shown in Table \ref{tab3}. Notably, EquiBoost-RS significantly outperforms other methods in generation quality.

\section{Visualizations}

\begin{figure}[h]
    \centering
    \includegraphics[width=0.46\textwidth]{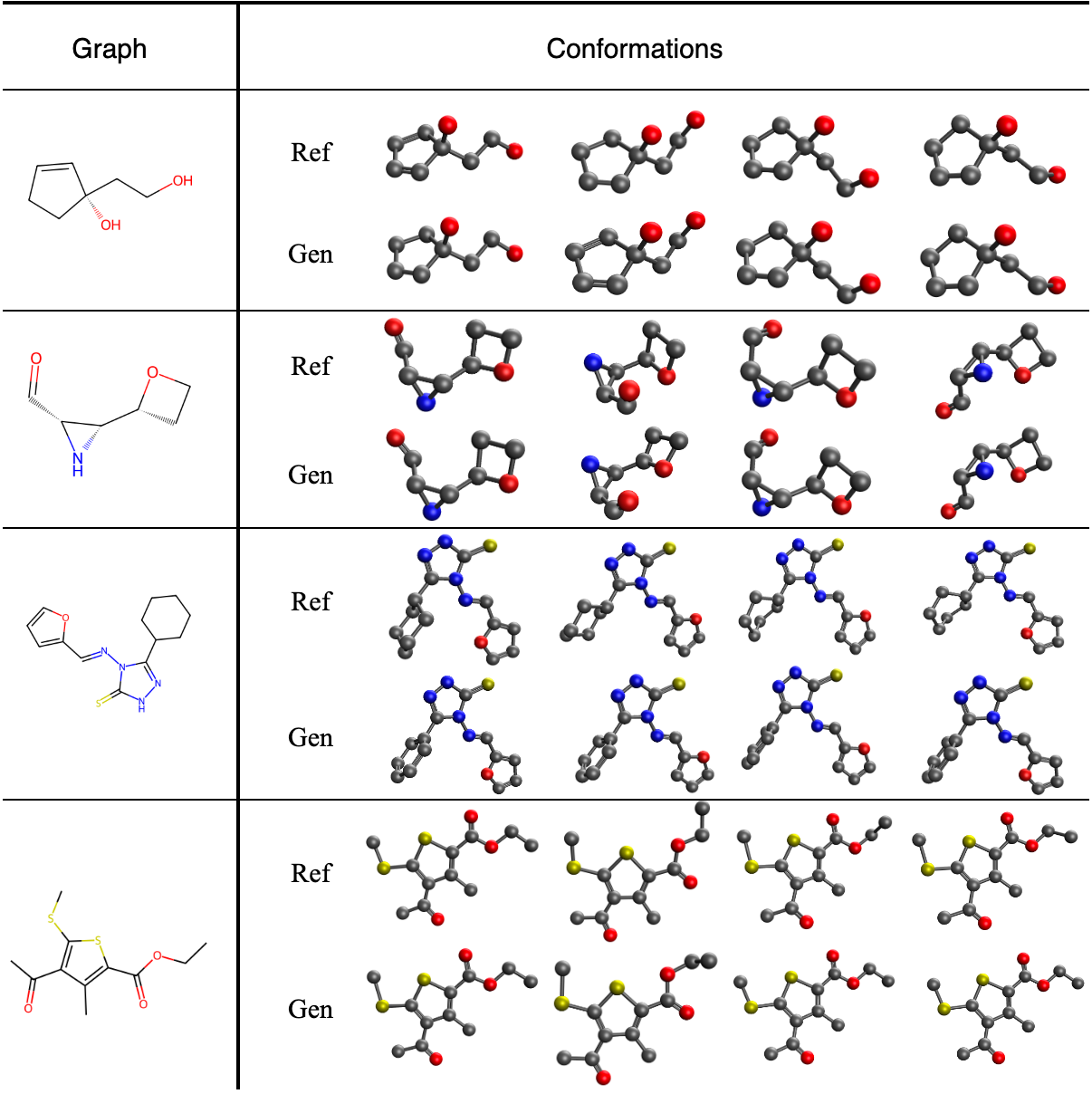}
    \caption{\textbf{Visualization of conformations generated by EquiBoost.}}
    \label{figS3}
\end{figure}

We provide some visualizations (Figure \ref{figS3}) of conformations generated by EquiBoost, using data from the test sets of GEOM-QM9 and GEOM-DRUGS.

\printbibliography[heading=subbibliography, title={References}, segment=1]
\end{document}